\documentclass[conference]{IEEEtran}
\IEEEoverridecommandlockouts
\usepackage{cite}
\usepackage{amsmath,amssymb,amsfonts}
\usepackage{algorithmic}
\usepackage{algorithm}
\usepackage{float}

\usepackage{graphicx}
\usepackage{textcomp}
\usepackage{bm}
\usepackage{xcolor}
\def\BibTeX{{\rm B\kern-.05em{\sc i\kern-.025em b}\kern-.08em
    T\kern-.1667em\lower.7ex\hbox{E}\kern-.125emX}}
\usepackage{hyperref}
\begin{document}

\title{Label Dependent Attention Model for Disease Risk Prediction Using Multimodal Electronic Health Records\\
}
\makeatletter

\author{
  \IEEEauthorblockN{Shuai NIU\textsuperscript{$^1$}, Qing YIN\textsuperscript{$^1$}, Yunya SONG\textsuperscript{$^2$}, Yike GUO\textsuperscript{$^1$}, Xian YANG\textsuperscript{$^{1,\ast}$ \thanks{* This is the corresponding author.}}}
  \IEEEauthorblockA{
  \textit{\textsuperscript{$^1$}The Department of Computer Science} \\
  \textit{\textsuperscript{$^2$}The Department of Journalism} \\
   \textit{Hong Kong Baptist University, Hong Kong, China}\\
    \{20483007, 21481326\}@life.hkbu.edu.hk, \{yunyasong, yikeguo, xianyang\}@hkbu.edu.hk}
}


 
  

\maketitle

\begin{abstract}
Disease risk prediction has attracted increasing attention in the field of modern healthcare, especially with the latest advances in artificial intelligence (AI). Electronic health records (EHRs), which contain heterogeneous patient information, are widely used in disease risk prediction tasks. One challenge of applying AI models for risk prediction lies in generating interpretable evidence to support the prediction results while retaining the prediction ability.
In order to address this problem, we propose the method of jointly embedding words and labels whereby attention modules learn the weights of words from medical notes according to their relevance to the names of risk prediction labels. 
This approach boosts interpretability by employing an attention mechanism and including the names of prediction tasks in the model. However, its application is only limited to the handling of textual inputs such as medical notes.
In this paper, we propose a label dependent attention model LDAM to 1) improve the interpretability by exploiting Clinical-BERT (a biomedical language model pre-trained on a large clinical corpus) to encode biomedically meaningful features and labels jointly; 2) extend the idea of joint embedding to the processing of time-series data, and develop a multi-modal learning framework for integrating heterogeneous information from medical notes and time-series health status indicators. 
To demonstrate our method, we apply LDAM to the MIMIC-III dataset to predict different disease risks. We evaluate our method both quantitatively and qualitatively. Specifically, the predictive power of LDAM will be shown, and case studies will be carried out to illustrate its interpretability.

\end{abstract}

\begin{IEEEkeywords}
label dependent attention model, disease risk prediction, Clinical-BERT, multimodal electronic health record analysis
\end{IEEEkeywords}

\section{Introduction}
Modern medicine aims at providing personalized healthcare to individual patients based on their clinical conditions. Artificial intelligence (AI) is being increasingly applied in electronic health records (EHRs) systems to improve data discovery and extraction and personalized clinical decision support.  Predictive models can be built from EHRs for a variety of important clinical problems and biomedical tasks, such as disease risk prediction \cite{cheng2016risk,harutyunyan2019multitask,choi2016doctor,mullenbach2018explainable}, patient ICU staying time prediction \cite{xu2018raim,harutyunyan2019multitask}, mortality prediction \cite{xu2018raim,harutyunyan2019multitask}, phenotype diseases prediction\cite{baytas2017patient,harutyunyan2019multitask}, and etc.
A typical EHR has various modalities, comprising complex facts and figures from  electrocardiogram (ECG) waveform, medical notes, laboratory testing, treatments, medications, diagnoses, and demographic information. Common data types of these modalities include unstructured texts, time-series signals, and code sequences. A number of studies have attempted to extract information from different modalities. 

 For example, to handle the time-series data,  \cite{lauritsen2020explainable}  adopted temporal convolutional networks (TCNs) for acute critical disease prediction. For the unstructured text data, \cite{mullenbach2018explainable,wang2018joint} developed natural language processing (NLP) based methods to embed textual data into vectors with which recurrent neural network (RNN) or attention-based models generate predictive results. For the sequence data (e.g., ICD codes \cite{ma2018kame,qiao2019mnn}), \cite{shang2019gamenet} proposed augmented memory networks for drug recommendations.

In this paper, we look into an important application of EHRs, i.e., disease risk prediction. In recent years, deep learning-based methods, such as attention-based RNN \cite{choi2016retain,xu2018raim}, convolutional neural networks (CNN) \cite{che2017exploiting,che2017boosting,cheng2016risk} and Transformer \cite{luo2020hitanet}, are increasingly applied to predict disease risks with EHRs. One challenge confronting the application of deep learning in the medical field is how to generate interpretable results while retaining predictive power. To cope with this problem, \cite{ma2018risk} integrated the prior medical rules (e.g., high blood pressure is a known risk factor for heart failure) into CNN and RNN models for risk prediction tasks. While this approach requires collecting medical rules for various diseases, there is a lack of prior knowledge for rare diseases. Apart from directly incorporating knowledge graphs into the models, the attention mechanism has been used to achieve model interpretability. For instance, \cite{xu2018raim} proposed a multi-channel attention model to analyze continuous ECG data, and this attention mechanism was guided by the time-series clinical observations.

Notably, we are interested in utilizing attention-based methods to achieve interpretability.  For the purpose of extracting information from the medical notes and assigning higher attention weights to words that are clinically more relevant to the prediction tasks, methods like CAML \cite{mullenbach2018explainable} and LEAM \cite{wang2018joint} were used to create a joint embedding of the medical notes and labels. The prediction performance was improved by making full use of label information for text classification. At the same time, these methods generated interpretable results by applying the cross-attention weights between embeddings of words and labels. \cite{xie2019ehr} further extended this line of research by using the knowledge graph to encode the names of labels. As labels of the prediction task in \cite{xie2019ehr} were ICD codes, the knowledge graph containing relationships among labels can be easily obtained. For our risk prediction tasks or other general prediction tasks, the labels of prediction tasks cannot directly form into a knowledge graph. As such, instead of using a knowledge graph, we apply a pre-trained biomedical language model, Clinical-BERT \cite{alsentzer2019publicly}, to create a joint embedding of task labels and  medical notes such that knowledge learned from a sizeable biomedical corpus can naturally be incorporated into the model. Labels and words in medical notes with similar meanings will be assigned similar embedding vectors by Clinical-BERT.

We propose a label-dependent attention model, LDAM, for disease risk prediction tasks by using 1) Clinical-BERT to jointly embed labels and features in the prediction model, and 2) a cross-attention mechanism to automatically select features relevant to task labels. Different from the previous work  \cite{mullenbach2018explainable,wang2018joint,xie2019ehr} that only use features such as words from medical notes, we adopt LDAM that handles time-series features as well as textual features. For time-series features (e.g., time-series health status indicators), we embed the feature names and assign different weights to different time signals based on the cross-attention between names of features and task labels. To the best of our knowledge, our work is the first of its kind to make full use of meaningful biomedical names of time-series features for predictive model construction. As both time-series features and textual features contain helpful information for disease risk prediction, we integrate information learned from both modalities to make the final risk prediction.
Our contributions can be summarized as follows:
\begin{itemize}
\item To generate interpretable results, we adopt a joint label and feature embedding model, which is based on a cross-attention mechanism that automatically selects features relevant to the prediction tasks. Clinical-BERT, a large language model pre-trained on a large clinical corpus, is used to generate biomedically meaningful embeddings.
\item We extend the usage of the joint label and feature embedding to time-series features. A multimodal learning framework based on fusing encoded information from time-series and textual features is developed.
\item To demonstrate our model, we apply our model to a collection of EHRs from the MIMIC-III database \cite{johnson2016mimic} and evaluate the performance in both quantitative and qualitative ways.  
\end{itemize}

We organize the rest of paper as follows. Section II discusses related work. Section III gives detailed descriptions of our LDAM model.
 Section IV shows experimental results and case studies. The conclusion is given in Section V.

\section{Related Work}
\subsection{Label Embedding in Predictive Modelling}
Label embedding has proved to work in a variety of areas and applications. In the area of computer vision (CV), \cite{akata2015label} proposed attribute label embedding (ALE) for zero-shot image classification by using images and texts. \cite{rodriguez2013label} embedded text images and words into a common latent space, and recognized the text image from the latest word embedding.  \cite{weston2010large} built a scalable model for image annotation by jointly embedding images and annotations. All of these works achieve the state-of-the-art performance in their tasks. 

The label embedding is also popular in the natural language processing (NLP) domain. \cite{zhang2017multi} converted both text and annotation into semantic vectors so that the classification task becomes a vector matching task.  \cite{mullenbach2018explainable} presented CAML and Deep CAML to predict ICD-9 codes from EHRs, which was comprised of CNN and attention mechanism. The attention was calculated using the  embeddings of text and labels. The difference between CAML and Deep CAML was that CAML used word-level label embedding and deep CAML used semantic label embedding. \cite{wang2018joint} followed CAML and  proposed LEAM for text classification and ICD-9 classification. They 
optimized the structure of CAML  by jointly learning text and label embeddings into the same latent space. 

Considering CAML and LEAM, the label embedding is not only valid to improve the prediction accuracy but can also generate more interpretative results. However, to the best of our knowledge, there is little work applying label embedding on time-series and multi-modal learning-related tasks.

\subsection{Attention Based Explainable Model Construction}
Deep learning models are usually data-driven. In recent years, due to the availability of big data, a variety of prediction tasks such as recognition and detection can be performed with high accuracy. Still, deep learning is criticized  as an intransparent black box and lacks interpretability. In the healthcare domain, accurate and interpretable results are crucial for clinicians to make clinical decisions. The attention mechanism is one of the most useful methods to improve the interpretability of deep learning models, which has been broadly adopted in the healthcare domain. 

For example, \cite{xu2018raim} proposed RAIM by using the self-attention  and guided-attention mechanisms, which used clinical events as the prior knowledge to regulate the continuous health status data. 
\cite{qiao2019mnn} Moreover, \cite{ma2017dipole} used the attentional bidirectional RNN.
GRAM \cite{choi2017gram} and KAME \cite{ma2018kame} employed the attention mechanism  to integrate information from the medical ontology into deep learning models, thus alleviating the problem of inadequate decision evidence. 
\cite{mullenbach2018explainable,wang2018joint} applied the cross-attention mechanism between patients' textual discharge summary and descriptions of ICD codes for ICD classification. 
\cite{xie2019ehr} proposed a cross-attention model with the multi-scale feature attention and the structured knowledge graph propagation to assign disease codes for EHRs. COMPOSE \cite{gao2020compose} applied a cross-attention between the clinical eligibility criteria (EC) sentence embedding and the taxonomy guided multi-granularity medical concept embedding to patient trials matching. 


\subsection{Biomedical Pre-trained Language Model in Biomedical Text Mining Tasks}
Transfer learning, which fine-tunes a large-scale language model pre-trained from a big corpus  for a new task, can  speed up the training process and introduce the domain-specific knowledge.
Clinical-BERT \cite{rasmy2020med} was trained on the MIMIC-III dataset with the aim of facilitating common  clinical AI tasks. \cite{rasmy2020med} fine-tuned Retain\cite{choi2016retain} and Clinical-BERT together to achieve a 3.3\% increase in the AUC score of the prediction task. 
Moreover, \cite{gao2020compose} proposed COMPOSE based on Clinical-Bert \cite{huang2019clinicalbert} to generate the embedding of clinical trial EC, which is then used as a query vector of the cross-attention module. 
For our proposed model LDAM, we apply Clinical-BERT to embed medical notes, names of time-series features, and names of disease risk labels to facilitate disease risk prediction.

\begin{table}[htbp]
\caption{ Notations and Descriptions }

\begin{center}

\begin{tabular}{|c|c|}
\hline
Notation & Description\\
\hline
$V$ &  The vocabulary \\
\hline
$|V|$ &  The  size  of vocabulary \\
\hline
$N_S$  & The number of time-series health status indicators\\
\hline
$N_Y$  & The number of disease risk labels\\
\hline
$\mathcal{T}$  & The number of time steps for health status indicators \\
\hline

$L$ & The number of tokens in a medical note\\
\hline
$D$ &   The size of embedding layer\\
\hline
$F$ &   The size of $\bm{z}^M$ and $\bm{z}^S$ \\
\hline
$\bm{X}$& Multimodal EHRs data\\
\hline
$\bm{S}$  & Numerical time-series data from health status indicators \\
\hline
$\bm{M}$  &  Medical notes from the discharge summary \\
\hline
$\mathcal{L}_Y$ &  The names of disease risk labels \\ 
\hline
$\mathcal{L}_S$ & The  names  of time-series health  status  indicators\\
\hline

$\bm{ y}$ & Disease risk labels \\
\hline

$\bm{\hat y}$ & The prediction of disease risk labels \\
\hline

$\bm{E}^M$ & The embedding  matrix of $\bm{M}$,$\in \mathbb{R}^{D \times L}$ \\
\hline
$\bm{E}^Y$ & The embedding  matrix of $\mathcal{L}_Y $,$\in \mathbb{R}^{D \times N_Y}$  \\
\hline
$\bm{E}^S$ & The embedding  matrix of $\mathcal{L}_S$,$ \in \mathbb{R}^{D \times N_S}$ \\
\hline

$\bm{H^c}$ & The output from channel-wise BiGRU\\
\hline
$\bm{H}$ & The output from BiGRU\\
\hline

$\bm{\beta}$ &  The cross-attention vector for medical notes \\
\hline
$\bm{\alpha}$ &  The cross-attention vector for time-series data\\
\hline
$\bm{z}^S$ &  The  weighted output of the time-series analysis module \\
\hline
$\bm{z}^M$ &  The weighted output of the text analysis module \\
\hline

$f_0$ &   Clinical-BERT encoder \\
\hline
$f_1$,$f_6$,$f_7$ &  Fully connected layer \\
\hline
$f_2$ &  1D CNN layer \\
\hline
$f_3$ &   Channel-wise  BiGRU layer\\
\hline
$f_4$ &   BiGRU layer\\
\hline
$f_5$ &  Cross-attention layer for numeric time-series data \\
\hline

\end{tabular}

\label{table:notations}
\end{center}
\end{table}

\section{Methods}

\subsection{Problem Definition and Notations}
As discussed in previous sections, the EHRs data contain heterogeneous information from multiple modalities.
Integrating information from multimodal data would help attain a comprehensive profile of patients and thus make the prediction model more robust.
In this work, we focus on using two typical modalities of EHRs as our model inputs, which contain the time-series health status indicators  and medical notes, to predict multiple health disease risks  (risks are defined in \cite{harutyunyan2019multitask}).
We define the input as $\bm{X}=\{ \bm{M}, \bm{S}, \mathcal{L}_S, \mathcal{L}_Y \}$. Let $\bm{M} = \{\bm{m}_1,...,\bm{m}_L\}$ represent the text sequence from medical notes, where each word $\bm{m}_l \in \mathbb{R}^{|V|}$ is a one hot vector and $|V|$ is the size of vocabulary $V$. 
$\bm{S}\in \mathbb{R}^{N_S \times \mathcal{T}}$ represents the numerical time-series data from health status indicators, where $N_S$ is the number of indicators, and $\mathcal{T}$ is the number of time steps.
The names of health status indicators are represented by $\mathcal{L}_S =\{\bm{l}_1,...,\bm{l}_{N_S}\}$,  in which the $v$th element of the name vector $\bm{l}_n\in \mathbb{R}^{|V|}$ equals 1 if the $v$th word in $V$ appears in the name of the $n$th indicator (otherwise it will be 0). The names of disease risk labels are represented as $\mathcal{L}_Y =\{\bm{l}_1,...,\bm{l}_{N_Y}\}$ with the definition same as $\mathcal{L}_S$.

Our problem can be defined as follows. Given a training set, where each sample is represented as $(\bm{X}, \bm{y})$.
 $\bm{X} \in \mathcal{X}$ contains multimodal information from an EHR as defined above, and $\bm{y} \in \mathcal{Y}$ is its corresponding disease risk labels. The goal for disease risk prediction is to learn a prediction function $f: \mathcal{X} \rightarrow \mathcal{Y}$ by minimizing the prediction loss.
The main notations used in this section are described in Table \ref{table:notations}.

\subsection{Overview}
\begin{figure*}[htbp]
\centerline{\includegraphics[scale=0.70]{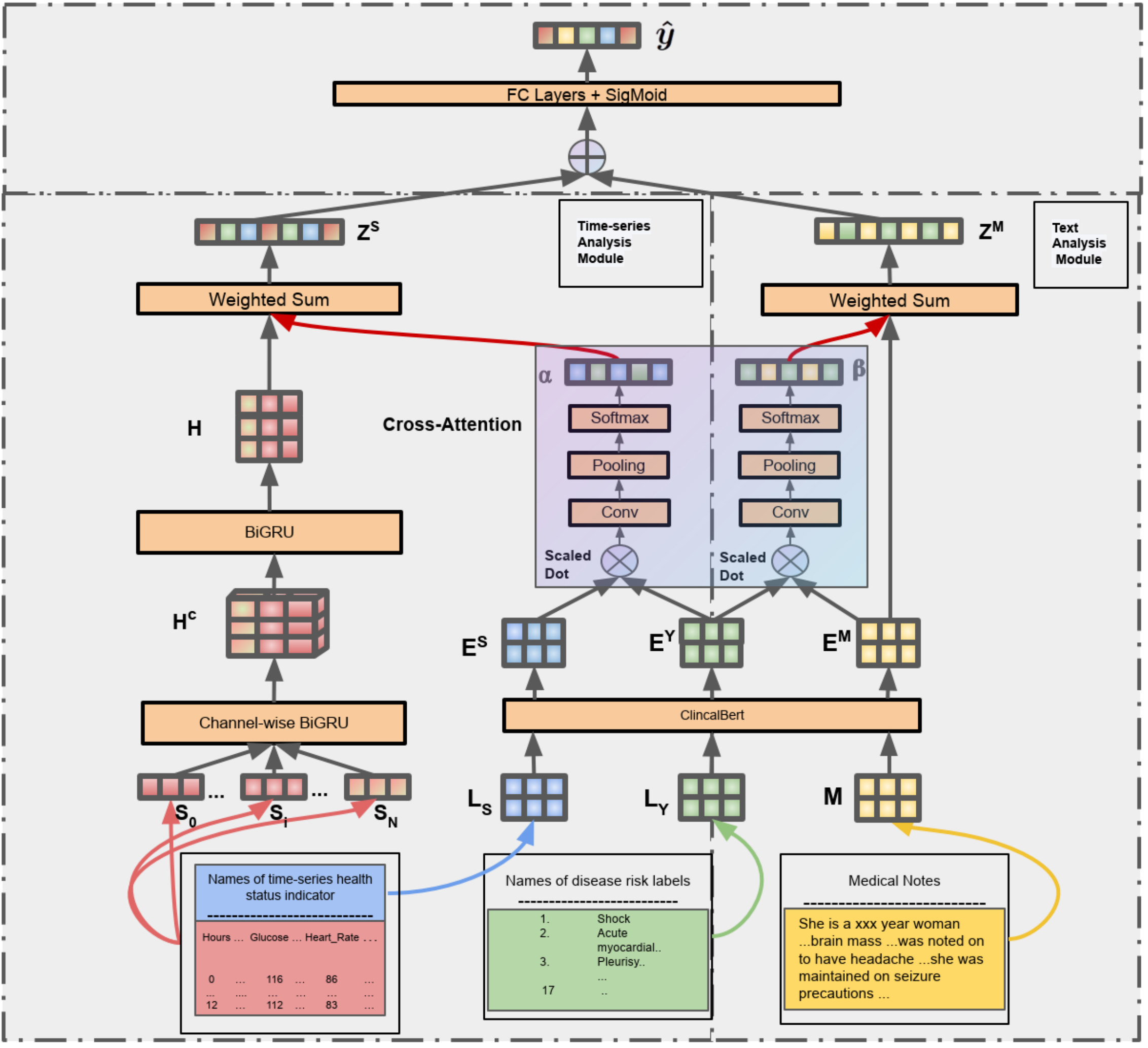}}
\caption{ The structure of LDAM. Our model LDAM is composed of two modules: the text analysis module (on the right) and the time-series analysis module (on the left). The text analysis module takes the medical notes and names of disease risk labels as the inputs. The time-series analysis module takes the signals of time-series health status indicators, names of time-series health status indicators, and  names of disease risk labels as the inputs. The core part of these two modules is the cross-attention layer to generate aggregated encoding vectors. The outputs of these two modules are integrated to make the final prediction. }
\label{fig:LDAM}
\end{figure*}

The details of our proposed model, LDAM, is illustrated in Fig.~\ref{fig:LDAM}.
The embedding layer, Clinical-BERT, first encodes the medical notes $\bm{M}$, names of health status indicators $\mathcal{L}_S$, and names of disease risk labels $\mathcal{L}_Y$ into embedding matrices $\bm{E}^M$, $\bm{E}^S$, and $\bm{E}^Y$, respectively. 
Then the text analysis module converts the information from medical notes $\bm{E}^M$ based on a cross-attention layer to assign weights to words in the medical notes in relevance to the names of disease risk labels $\bm{E}^Y$.
 On the other hand, the time-series analysis module encodes signals from time-series health indicators $\bm{S}$ and aggregate multiple signals based on the cross-attention between the embeddings of names of health indicators $\bm{E}^S$ and disease risk labels $\bm{E}^Y$. The outputs of the text analysis module $\bm{z}^M$ and the time-series analysis module $\bm{z}^S$ are integrated together to make the prediction of risks $\bm{y}$.

\subsection{Embedding Layer}
The text sequences from medical notes $\bm{M}$ are first passed through an embedding layer $f_0:\bm{M} \rightarrow \bm{E}^M$, in which $\bm{E}^M \in \mathbb{R}^{D \times L}$ is the embedding matrix and $D$ is the embedding size. 
Similarly, the names of disease risk labels $\mathcal{L}_Y$ are passed through $f_0$ to get its embedding matrix  $\bm{E}^Y \in \mathbb{R}^{D \times N_Y}$.
The names of time-series health status indicators $\mathcal{L}_S$ are also passed through $f_0$ to get its embedding matrix  $\bm{E}^S \in \mathbb{R}^{D \times N_S}$.
In our model, $f_0$ is implemented by Clinical-BERT. 
 Clinical-BERT was trained on a big clinical corpus \cite{johnson2016mimic} with the aim of facilitating various downstream disease-prediction tasks. Thus, Clinical-BERT is a perfect choice of our  embedding layer. To get embedding matrices $\bm{E}^Y$ and $\bm{E}^S$, we use the embedding vector associated with  the [CLS] token from Clinical-BERT's output to represent the embedding of a sequence (i.e., a name containing multiple medical words). On the other hand,  Clinical-BERT encodes medical notes and returns $L$ embedding vectors  for all $L$ words to generate $\bm{E}^M$.

\subsection{Text Analysis Module}
We adopt the idea in LEAM\cite{wang2018joint} to build up a text analysis module, which is illustrated in Fig.~\ref{fig:LDAM}.
$\bm{E}^M$ and  $\bm{E}^Y$ are used to compute the cross-attention score matrix $\bm{\beta}$, which can help to highlight important words from the text. We first apply a fully connection layer $f_1$ to reduce the dimension of $\bm{E}^M$ and  $\bm{E}^Y$ from  $D$ to $F$.
The outputs of $f_1$ are then fed into a cross-attention layer $f_2$ as follows:
\begin{equation}
\bm{G} = ScaledDot(f_1(\bm{E}^M), f_1(\bm{E}^Y)) =  \frac{({f_1(\bm{E}^M)})^T* f_1(\bm{E}^Y)}{\sqrt{F}}\label{eq:1}
\end{equation}
where the $T$ is the transpose operator, the cross-attention matrix $\bm{G}\in \mathbb{R}^{L \times N_Y}$ is the scaled-dot similarity and the operator $*$ is the matrix product.

To better capture the relative spatial information of consecutive words and improve the ability to extract implicit information from texts, we apply a one-dimensional (1D) CNN together with a max-pooling layer on the cross-attention matrix $\bm{G}$ as follows:

\begin{equation}
\bm{u} = f_2(\bm{G}) = MaxPool(ReLU(Conv(\bm{G},k_1,q)), k_2)\label{eq:2}
\end{equation}
where $Conv$ is the 1D CNN layer, $MaxPool$ is the max-pooling layer, $ReLU$ is the nonlinear activation layer, $k_1$ is the kernel width (N-Gram) of CNN, $q$ is the padding size of CNN (set to `same padding' in our implementation), and $k_2$ is the kernel width of MP. 
The output vector $\bm{u} \in \mathbb{R}^{L}$ is then normalized as follows:
\begin{equation}
\bm{\beta} = SoftMax(\bm{u})\label{eq:3}
\end{equation}
where $\bm{\beta} \in \mathbb{R}^{L} $ and  the $l$-th element of SoftMax is $\beta_l = \frac{exp(u_l)}{\sum ^L_{l=1}exp(u_l)}$.

Using $\bm{\beta}$, we can obtain the weighted representation of the text data, which is expressed as follows:

\begin{equation}
\bm{z}^M = f_1(\sum_{l=1}^{L} \bm{E}_l^M \bm{\beta}_l)\label{eq:4}
\end{equation}
where $\bm{E}_l^M \in \mathbb{R}^D$ is the $l$th column of $\bm{E}^M$, and $\bm{\beta}_l$ is the $l$th element of $\bm{\beta}$,
and $\bm{z}^M \in \mathbb{R}^{F}$ is the output of the text analysis module.


\subsection{Time-series Analysis Module}
The time-series analysis module shown in Fig.~\ref{fig:LDAM} takes $\bm{S}$, $\bm{E}^S$, and $\bm{E}^Y$ as the inputs.
The channel-wise bidirectional gated recurrent unit (BiGRU) \cite{chung2014empirical} is used as the basic building block for encoding the time-series data $\bm{S}$. A cross-attention layer measures the similarity between $\bm{E}^S$ and $\bm{E}^Y$ for generating the weighted representation vector $\bm{z}^S$ as the module output.

\textbf{Time-series Feature Extraction Layer}: We follow the similar approach in \cite{harutyunyan2019multitask} to apply a channel-wise BiGRU to extract local temporal features. For the $i$th health indicator (i.e., channel), we have
\begin{equation}
\bm{h}_i^c = f_3(\bm{S}_i) = BiGRU(\bm{S}_i), \forall i \in \{1,...,N_s\}\label{eq:5}
\end{equation}
where $\bm{S}_i \in \mathbb{R}^\mathcal{T}$ is the $i$th row of $\bm{S}$, and 
$\bm{h}_i^c \in \mathbb{R}^{\mathcal{T} \times F}$ is the output of BiGRU for $\bm{S}_i$. Let 
$\bm{H}^c \in \mathbb{R}^{N_S \times \mathcal{T} \times F}$ denote the outputs of all channels:
\begin{equation}
\bm{H}^c = [\bm{h}^c_1,..,\bm{h}^c_{N_S}]^T\label{eq:6}
\end{equation}
We then feed $\bm{H}^c$ into another BiGRU to integrate information and obtain the last hidden state across different channels:

\begin{equation}
\bm{H} = f_4(\bm{H}^c) = BiGRU(\bm{H}^c)\label{eq:7}
\end{equation}

where $\bm{H} \in \mathbb{R}^{N_S \times F}$ contains the information extracted from time-series numerical data.

\textbf{Cross-Attention Layer}: As shown in Fig.~\ref{fig:LDAM}, we also adopt the cross-attention and CNN layers to generate the cross-attention vector using $\bm{E}^S$ and $\bm{E}^Y$:



\begin{equation}
\begin{aligned}
\bm{\alpha}& = f_5(\bm{E}^S,\bm{E}^Y)\\
& =  SoftMax(MaxPool(ReLU(Conv(ScaledDot \\
& (f_1(\bm{E}^S),f_1(\bm{E}^Y)),k_1,q), k_2)))\label{eq:8}
\end{aligned}
\end{equation}
where $\bm{\alpha}\in \mathbb{R}^{N_S}$ is the attention vector.
Using $\bm{\alpha}$, we can get the weighted output $\bm{z}^S \in \mathbb{R}^F$ as:
\begin{equation}
\bm{z}^S = \sum_{i=1}^{N_s} \bm{H}_i  \bm{\alpha}_i\label{eq:9}
\end{equation}
where $\bm{H}_i \in \mathbb{R}^F$ is the $i$th row of $\bm{H}$ and $\bm{\alpha}_i$ is the $i$th element in $\bm{\alpha}$.

\subsection{Output Layer and Model Training}
\textbf{Output Layer}:
After we have obtained $\bm{z}^M$ and $\bm{z}^S$, we integrate them together into one vector through 
two fully connected layers $f_6$ and $f_7$ as follows:
\begin{equation}
\bm{z} =  f_7(f_6(\bm{z}^M \oplus \bm{z}^S))\label{eq:10}
\end{equation}
where the $\oplus$ is the concatenation operator. $\bm{z} \in \mathbb{R}^{N_Y}$ is used to make the final prediction via:

\begin{equation}
\bm{\hat \bm{y}} = Sigmoid(\bm{z})\label{eq:11}
\end{equation}
where the $j$th element in $\bm{\hat \bm{y}} \in \mathbb{R}^{N_Y}$ is $\hat y_j=\frac{1}{(1+exp(-z_j)}$.

\textbf{Training Objective}
In LDAM, the loss for model training is defined as follows:

\begin{equation}
\begin{aligned}
Loss& = -\frac{1}{N_Y} \sum_{j=1}^{N_Y} (y_j \cdot \log(\hat y_j)) + (1-y_j)\cdot \log(1-\hat {y}_j))\\
& -\frac{1}{N_Y} \sum_{j=1}^{N_Y} \bm{p}_j  \cdot \log(  SoftMax(f_7(f_1(\bm{E}^Y_j)))),
\label{eq:12}
\end{aligned}
\end{equation}
where $y_j \in \{0,1\}$ indicates the presence of the $j$th disease risk, and  $\bm{p}_j$ is a one hot vector whose $j$th element equals 1.
The first part of this loss is the classification loss of our multilabel disease risk prediction problem, while the second part is to distinguish different embeddings of the names of disease risk labels.
Our training objective is minimizing the loss over all data samples by the Adam\cite{kingma2014adam} optimizer. The whole training procedure of LDAM is shown in Algorithm 1.

\begin{algorithm}
    \caption{Label Dependent Attention Model}
    \label{alg:1}
    \begin{algorithmic}
        \STATE {\textbf{Input}}
            \STATE { Given a training set, where each sample is represented as $(\bm{X}, \bm{y})$. $\bm{X} \in \mathcal{X}$ contains multimodal information from an EHR and $\bm{y} \in \mathcal{Y}$ is its corresponding disease risk labels.}

      \STATE {\textbf{Training}}
            \STATE {\quad Initialize model parameters.}
            \FOR{iteration $i$ =1,2,... }
                \STATE {Select a minibatch of the training dataset.}
                \FOR{each EHR in the minibatch}
                \STATE {Generate the embeddings of medical notes, names of time-series disease risk labels, and names of health status indicators (i.e., $\bm{E}^M$, $\bm{E}^Y$, and $\bm{E}^S$) using $f_0$.}\STATE{Generate $\bm{z}^M$ from the text analysis module using
                $\bm{E}^M$ and $\bm{E}^Y$  based on \eqref{eq:1}\eqref{eq:2}\eqref{eq:3}\eqref{eq:4}.}
                \STATE{Generate $\bm{z}^S$ from the time-series analysis module using $\bm{S}$ (numerical time-series data from health status indicators), $\bm{E}^S$, and $\bm{E}^Y$ \eqref{eq:5}\eqref{eq:6}\eqref{eq:7}\eqref{eq:8}\eqref{eq:9} .}
                \STATE{Get the prediction results $\bm{\hat y}$ using $\bm{z}^{m}$ and $\bm{z}^{S}$ based on \eqref{eq:10}\eqref{eq:11}.}
                \ENDFOR
               \STATE{Update parameters by minimizing the loss (defined in \eqref{eq:12}) over all samples in the minibatch.}
          
            \ENDFOR

    \end{algorithmic}
\end{algorithm}

\section{Experiments}

\subsection{Dataset}

We work on a publicly accessible dataset MIMIC-III to demonstrate our proposed model. MIMIC-III is widely used to evaluate  different prediction models such as CAML, LEAM, and MLB \cite{harutyunyan2019multitask}.
For illustration purposes, we primarily use two types of information for risk prediction, i.e.,  17 time-series health status indicators are drawn from patient laboratory testing results (e.g., capillary refill rate, fraction inspired oxygen, heart rate) and medical notes from the discharge summary of a brief hospital course. 
We use 25 different disease risk labels defined in \cite{harutyunyan2019multitask} as our prediction tasks which cover multiple types of risks such as acute and unspecified renal failure, acute myocardial infarction, chronic kidney disease, hypertension with complications, cardiac dysrhythmias, diabetes mellitus with complications, etc.

To pre-process the medical notes, we  adopt the same approach as in CAML to remove the nonalphabetic characters \& stop-words and convert uppercase to lowercase letters. For the time-series health status indicators, we impute the missing values by the reference numbers used in MLB. 
There is a total of 31,485 unique EHRs selected from MIMIC-III with no missing information from medical notes, time-series health status indicators and disease risk labels.
To make a fair comparison, we follow the data splitting strategy in MLB to get 25,190 training and 6,294 testing samples (80\% for training and 20\% for validation).

\begin{table*}[!htbp]
\begin{center}

\caption{Performance of comparative methods}

\begin{tabular}{|c|c|c|c|c|c|c|}
\hline
&\multicolumn{6}{|c|}{\textbf{Evaluation Metrics}} \\
\cline{2-7} 
\textbf{Models} & Micro Precision & Macro Precision & Micro Recall & Macro Recall & Micro ROC AUC &  Macro ROC AUC \\
\hline
CG                            & 0.5973 & 0.3411 & 0.1932 & 0.1499 & 0.7870 & 0.7283 \\
\hline
CG+$\mathcal{S} $               & 0.6231 & 0.3148 & 0.1564 & 0.1209 & 0.7879 & 0.731 \\
\hline
CG+$\mathcal{C}$  & 0.6219 & 0.327 &  0.1681 & 0.1283 & 0.7889 & 0.7248  \\
\hline
LEAM                          & 0.8488 & 0.4955 & 0.2674 & 0.2130  & 0.8770 & 0.8582  \\
\hline
$\mathcal{B}+\mathcal{S}$              & 0.7563 & 0.6517 & 0.518 &0.4647 & 0.8849 & 0.8549 \\
\hline
$\mathcal{B}$+LEAM   & 0.7027 & 0.6522 & 0.61346 & 0.5633 & 0.888 & 0.8599 \\
\hline
LDAM             & 0.7906 & 0.6895 & 0.5235 & 0.4669 & 0.9036 & 0.8824 \\
\hline
\end{tabular}
\label{tab1:performance}
\end{center}

\end{table*}

\subsection{Comparative Methods and Implementation Details}
To demonstrate the effectiveness of introducing label dependent attention, we compare our LDAM model with different  methods, including  those only are focusing on the time-series data and also those are dealing with textual data. The model and hyperparameters of LDAM is available at  
 \footnote{\href{https://github.com/finnickniu/LDAM}{https://github.com/finnickniu/LDAM}}.
\subsubsection{Methods for time-series health status indicators}
\begin{itemize}
    \item \textbf{CG} (Channel-wise BiGRU): We apply a similar approach used in \cite{harutyunyan2019multitask} as one of our baseline models. It has  two BiGRU layers: one for processing individual multichannel features, and the other one for capturing a global temporal relation. Dimensions of these two layers equals 256. These two BiGRU layers are followed by two fully connected layers for  prediction, with the size of each layer are equal to 128 and 64, respectively. 
    \item \textbf{CG+$\bm{\mathcal{S}}$} (Channel-wise BiGRU + self-attention): We employ the self-attention mechanism \cite{vaswani2017attention} on the basis of CG. It is used to show whether the self-attention module without including information contained in labels can achieve better performance than CG. The Scaled-Dot and CNN with Softmax is used to generate attention weights.
    \item \textbf{CG+$\bm{\mathcal{C}}$} (Channel-wise BiGRU + cross-attention): Instead of using a self-attention mechanism, we adopt a cross-attention module.
    The inputs of the cross-attention layer are
    the embeddings of the names of time-series health status indicators and disease risk labels generated by Clinical-BERT. The dimension of embedding vectors is 768. Cross-attention is designed in the same way as in our LDAM model. Here, CG+$\mathcal{C}$ is to show the effectiveness of incorporating the label dependent idea for analyzing time-series data. 
\end{itemize}
\subsubsection{Methods for medical notes}
\begin{itemize}
    \item \textbf{LEAM}: LEAM is one of state-of-the-art deep learning model, particularly designed for ICD-9 code prediction using textual information from medical notes. The structure of LEAM consists of a joint word, label embedding layer, a cross-attention module, and multiple fully connected layers. Different from our model, the embedding layers are randomly initialized and updated through end-to-end training. We choose the default settings of LEAM as implemented in \footnote{\href{https://github.com/guoyinwang/LEAM}{https://github.com/guoyinwang/LEAM}}.
    \item $\bm{\mathcal{B}+\mathcal{S}}$ (Clinical-BERT + self-attention): 
  Clinical-BERT is used as an encoder, while self-attention is used to generate a weighted feature representation. We use the default settings of Clinical-BERT as implemented in \footnote{\href{https://github.com/EmilyAlsentzer/clinicalBERT}{https://github.com/EmilyAlsentzer/clinicalBERT}}. For self-attention, we adopted the scaled-dot similarity (refer to \eqref{eq:1} in the method section) and N-Gram CNN with Softmax to generate attention weights. The attention weights will be applied for the weighted sum to represent the embedding of the whole documents.
    \item \textbf{$\bm{\mathcal{B}}$+\textbf{LEAM}} (Clinical-BERT + LEAM): As a comparative method of LEAM and  \textbf{$\mathcal{B}$+$\mathcal{S}$}, we embed medical notes and names of disease risk labels by Clinical-BERT and use  cross-attention instead of self-attention based on their embeddings. The cross-attention module is the same as the one in our LDAM model.
\end{itemize}
\subsubsection{Our method}
\begin{itemize}
    \item \textbf{LDAM}: This is the multimodal learning model proposed in this paper.  Medical notes, names of disease risk labels, and names of time-series health status indicators are embedded by Clinical-BERT. The  textual analysis module and the time-series analysis module encode textual and time-series data via the cross-attention mechanism.
\end{itemize}

The final output layer of all comparative models is a fully connected layer followed by the Sigmoid function.  The same structures of all comparative methods use the same set of hyperparameters. We choose  Adam\cite{kingma2014adam} as the optimizer for all models. Our model, LDAM, was trained on GPU Tesla V100S with Pytorch.

\subsection{Quantitative analysis}
The performance of all comparative models is evaluated using the following metrics: precision, recall, and ROC AUC (area under the curve). We calculate both micro- and macro-averages for these metrics. In our experiments, 25 different risk labels are predicted from patients' time-series health status indicators and medical notes. In this multi-class prediction setup, the macro-average treats all classes equally,  whereas the micro-average takes the number of each class into consideration.

Table \ref{tab1:performance} shows the results of all comparative methods, from which we have the following observations:
\begin{itemize}
    \item First, we can see that the methods only using time-series health status indicators as their input data (i.e., CG, CG+$\mathcal{S}$, and CG+$\mathcal{C}$) perform much worse than the other methods (i.e., LEAM, $\mathcal{B}+\mathcal{S}$, $\mathcal{B}$+LEAM, and LDAM) having incorporated medical notes for making predictions. This observation indicates that information contained in unstructured  medical notes are useful
   to reflect patient disease status.
    \item Secondly, for the time-series data, the cross-attention based  model CG+$\mathcal{C}$ achieves a slightly better performance than the self-attention based model CG+$\mathcal{S}$ in terms of all evaluation metrics.This implies that  label dependent attention can also be helpful for time-series data (not limited to textual data).
    
    \item Thirdly, for the text data, the cross-attention-based model without using Clinical-BERT (i.e., LEAM) has not been found to be better than $\mathcal{B}+\mathcal{S}$. 
    Especially, LEAM has much lower recall values than $\mathcal{B}+\mathcal{S}$. 
    This is because $\mathcal{B}+\mathcal{S}$ has exploited  Clinical-BERT to learn better embeddings of medical words. This observation demonstrates the power of the pre-trained language model. 
    
    \item Fourthly, for the text data, compared with LEAM, and $\mathcal{B}+\mathcal{S}$, $\mathcal{B}$+LEAM can achieve an increase in recall values and slightly higher ROC AUC values. This observation means that by using  Clinical-BERT as the embedding layer and adopting the label dependent attention, the chance of failing to predict risk would be significantly reduced.
    
    \item Finally,
    out of all comparative methods, our LDAM has achieved the best performance with the highest ROC AUC values. It indicates our multimodal model  can achieve better performance than the single modality model. 
\end{itemize}

\subsection{Qualitative analysis}

\subsubsection{Visualization of Embeddings}
Here, we would like to visualize the embeddings of three types of texts (i.e., medical notes, names of time-series health status indicators, and names of disease risk labels) to check whether semantically similar texts would have similar embeddings generated from LDAM.  

\begin{figure}[htbp]
\centerline{\includegraphics[scale=0.63]{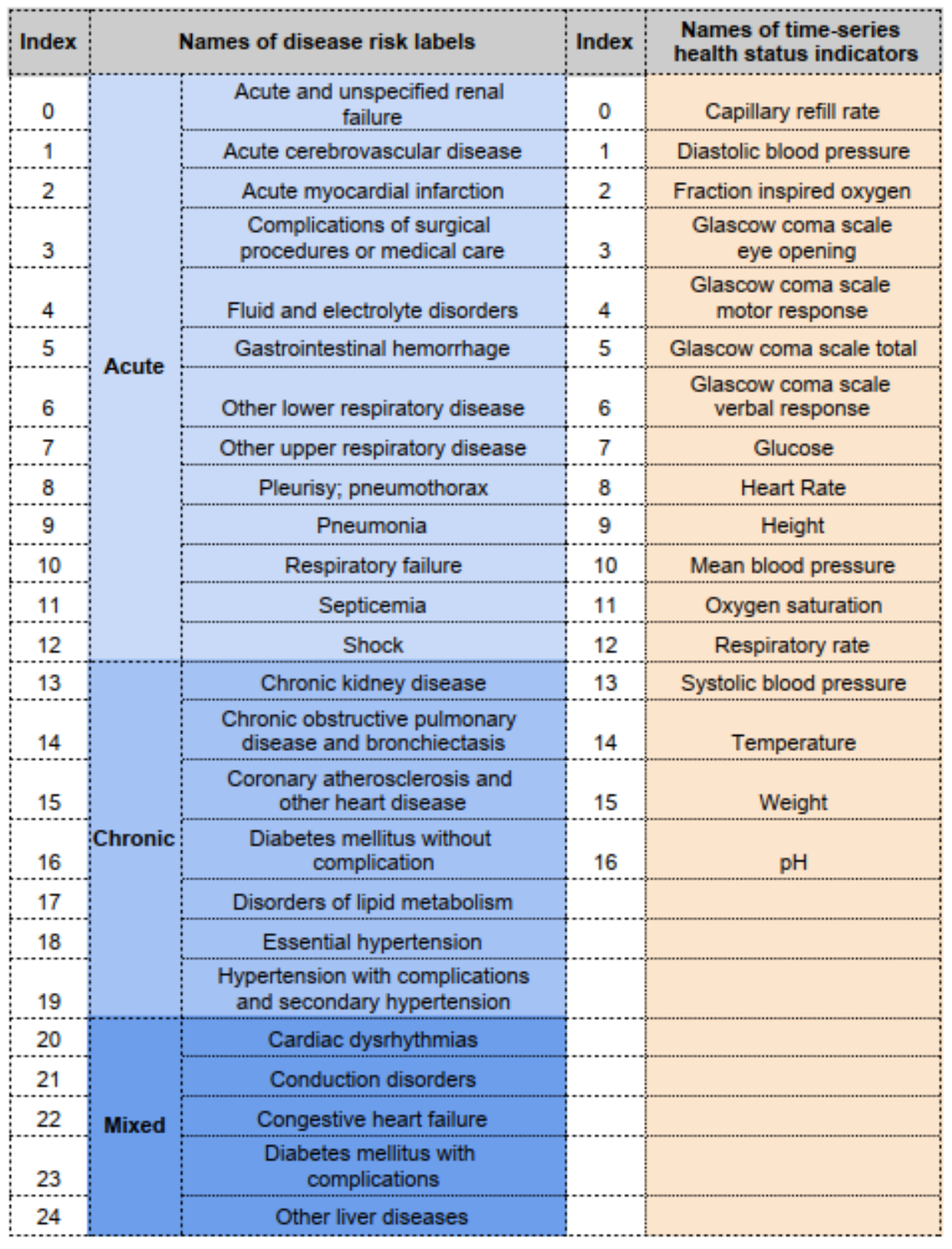}}
\caption{ Names of time-series health status indicators and disease risk labels.}
\label{fig:note}
\end{figure}

\begin{figure*}[htbp]
\centerline{\includegraphics[scale=0.80]{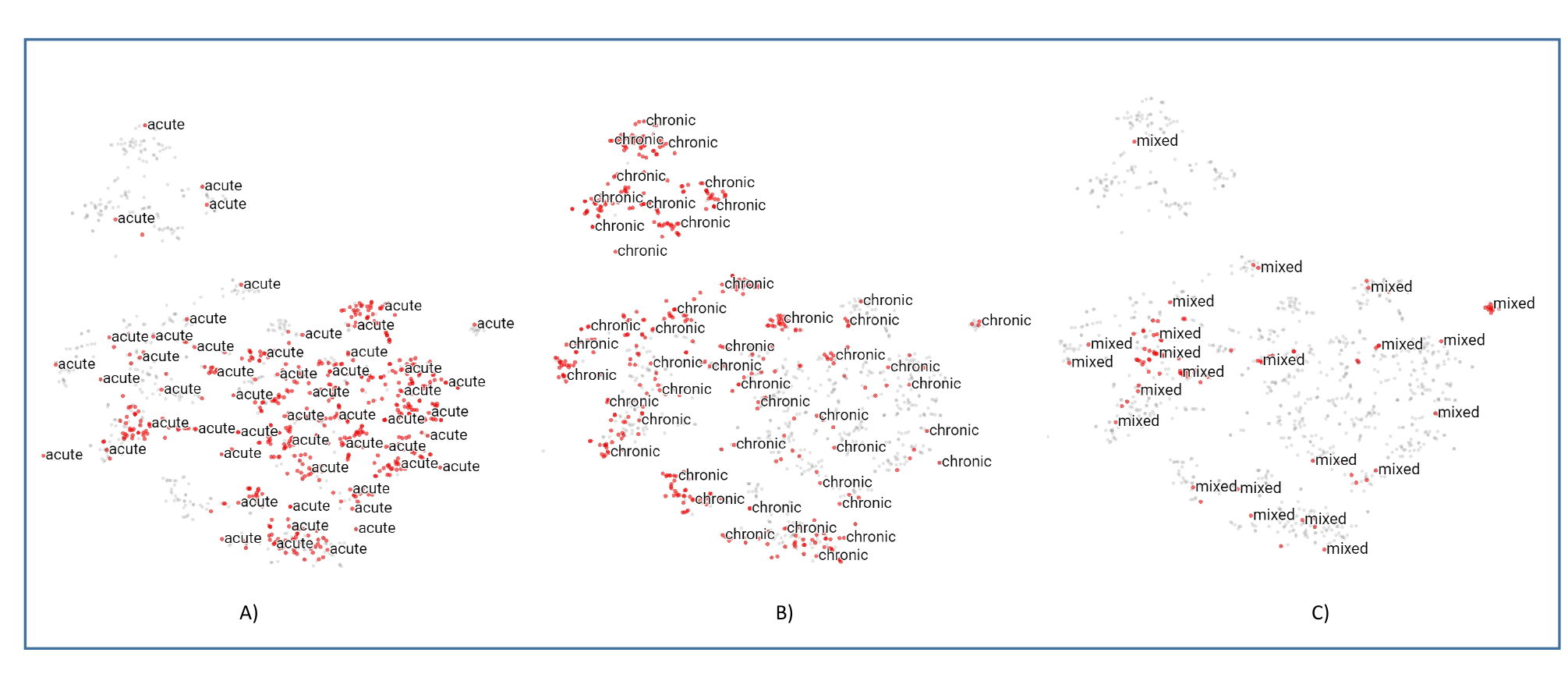}}
\caption{The t-SNE visualization of embeddings of medical notes from patients with different types of disease risk. Patients with acute, chronic and mixed disease risk are highlighted in A), B) and C), respectively.  }
\label{fig:3type}
\end{figure*}

Fig.~\ref{fig:note} gives detailed information on time-series health status indicators and  risk labels.
25 disease risk labels can be grouped into 3 main groups, which are acute disease risk, chronic disease risk, and mixed disease risk. 
There are in total 564
patients of acute type, 104  patients of mixed type, and 478  patients of chronic type from the testing datasets. Other patients who coherently have more than one risk type are not included in the groups of  Fig.~\ref{fig:note}. t-SNE \cite{van2008visualizing} is used for visualizing high-dimensional embedding vectors in low dimensional space.
Fig.~\ref{fig:3type} shows the 2D visualization of the embeddings of medical notes
where patients with acute, mixed, and chronic disease risk types are highlighted in subplots A), B), and C), respectively.
From Fig.~\ref{fig:3type} A), we can find that patients with acute disease risk are more likely to be clustered together. Patients with chronic disease risk are scattered all around the space in Fig.~\ref{fig:3type} B). However, we can still see some clusters in the upper part of this subplot mainly containing chronic patients. Similarly, patients with mixed disease risk cannot form into a single cluster in Fig.~\ref{fig:3type} C). These observations indicate the heterogeneous characteristics of chronic and mixed patients.

\begin{table*}[!htbp]
\caption{Interpretability of different models on medical notes. }

\begin{center}
\resizebox{520pt}{85pt}{

\begin{tabular}{|c|c|c|}
\hline
Patient &\textbf{Acute Risk Disease }  & Fluid and electrolyte disorders, Gastrointestinal hemorrhage \\
\hline
& &  ... erythema granularity \textbf{friability and congestion} in the \textbf{whole} stomach \textbf{compatible} with\textbf{ gastritis erythema} and congestion \\
P1 & \textbf{LEAM}&   in the \textbf{duodenal} bulb \textbf{compatible} with \textbf{duodenitis} otherwise\textbf{ normal egd to third part} of the\textbf{ duodenum } ...  \\
\hline

 & &  ... \textbf{erythema granularity} friability and\textbf{ congestion in the whole} stomach compatible \textbf{with} gastritis erythema and congestion \\
P1 &\textbf{$\mathcal{B}$+Leam} &   \textbf{in} the  duodenal bulb compatible with duodenitis \textbf{otherwise normal} egd to \textbf{third} part \textbf{of} the duodenum ...  \\
\hline

 &&  ...\textbf{ erythema granularity friability} and \textbf{congestion in the} whole \textbf{stomach} compatible \textbf{with gastritis erythema} and \textbf{congestion} \\
P1 &\textbf{LDAM} & \textbf{ in the duodenal bulb} compatible with\textbf{ duodenitis otherwise} normal \textbf{egd} to\textbf{ third part of the duodenum} ...  \\
\hline

&\textbf{Chronic Risk Disease } &Coronary atherosclerosis and other heart disease, Disorders of lipid metabolism, Essential hypertension, Cardiac dysrhythmias \\
\hline
 &&  ... a heavily \textbf{calcified} aorta and \textbf{coronaries} dr last name stitle was \textbf{consulted and }on \textbf{he} underwent \textbf{cabgx3} with \textbf{lima} lad\textbf{ svg pda}  \\
P2&\textbf{LEAM}&   and \textbf{om} cross clamp \textbf{time} was \textbf{minutes} and \textbf{total bypass time} was \textbf{minutes} he required a \textbf{urology} consult \textbf{intraoperatively} ...  \\
\hline

 && ...  a heavily calcified aorta and coronaries dr \textbf{last} name \textbf{stitle} \textbf{was consulted and on he underwent cabgx3} with lima lad svg pda \\
P2 &\textbf{$\mathcal{B}$+Leam}&    and om cross \textbf{clamp} time was minutes and total bypass time was minutes \textbf{he required} a urology consult \textbf{intraoperatively} ...  \\
\hline

 &&  ...\textbf{a} heavily\textbf{ calcified aorta and coronaries} dr last \textbf{name} stitle \textbf{was} consulted and \textbf{on} he underwent cabgx3 \textbf{with lima} lad \textbf{svg} pda \\
P2&\textbf{LDAM} &  and \textbf{om} cross \textbf{clamp time was} minutes and \textbf{total} bypass \textbf{time} was \textbf{minutes} he required \textbf{a urology consult intraoperatively}  ...  \\
\hline

&\textbf{Mixed Risk Disease } & Congestive heart failure \\
\hline
 &&   ... he had a \textbf{short period} of \textbf{hematuria} after \textbf{pulling} foley but \textbf{this} resolved ... he had \textbf{no} episodes \textbf{of} chest \textbf{pain} post \textbf{procedure}  \\
P3 &\textbf{LEAM} &   and \textbf{was discharged} with \textbf{follow} up with a \textbf{cardiologist} and \textbf{prescriptions} ...  \\
\hline

 &&  ...\textbf{ he had a short period of hematuria after pulling foley but this resolved ... he had no episodes of chest} pain post \textbf{procedure} \\
P3 &\textbf{$\mathcal{B}$+Leam} &  and \textbf{was discharged with follow} up with a cardiologist and prescriptions ...  \\
\hline

 & &  ... \textbf{he had a short period} of \textbf{hematuria} after \textbf{pulling foley} but this \textbf{resolved} ... he \textbf{had no episodes} of chest \textbf{pain} \textbf{post procedure} \\
P3 &\textbf{LDAM} & and \textbf{was} discharged with follow \textbf{up with} a \textbf{cardiologist} and \textbf{prescriptions}  ...  \\
\hline
\end{tabular}
}
\label{table:3Patient}
\end{center}
\end{table*}

\begin{figure}[htbp]
\centerline{\includegraphics[scale=0.70]{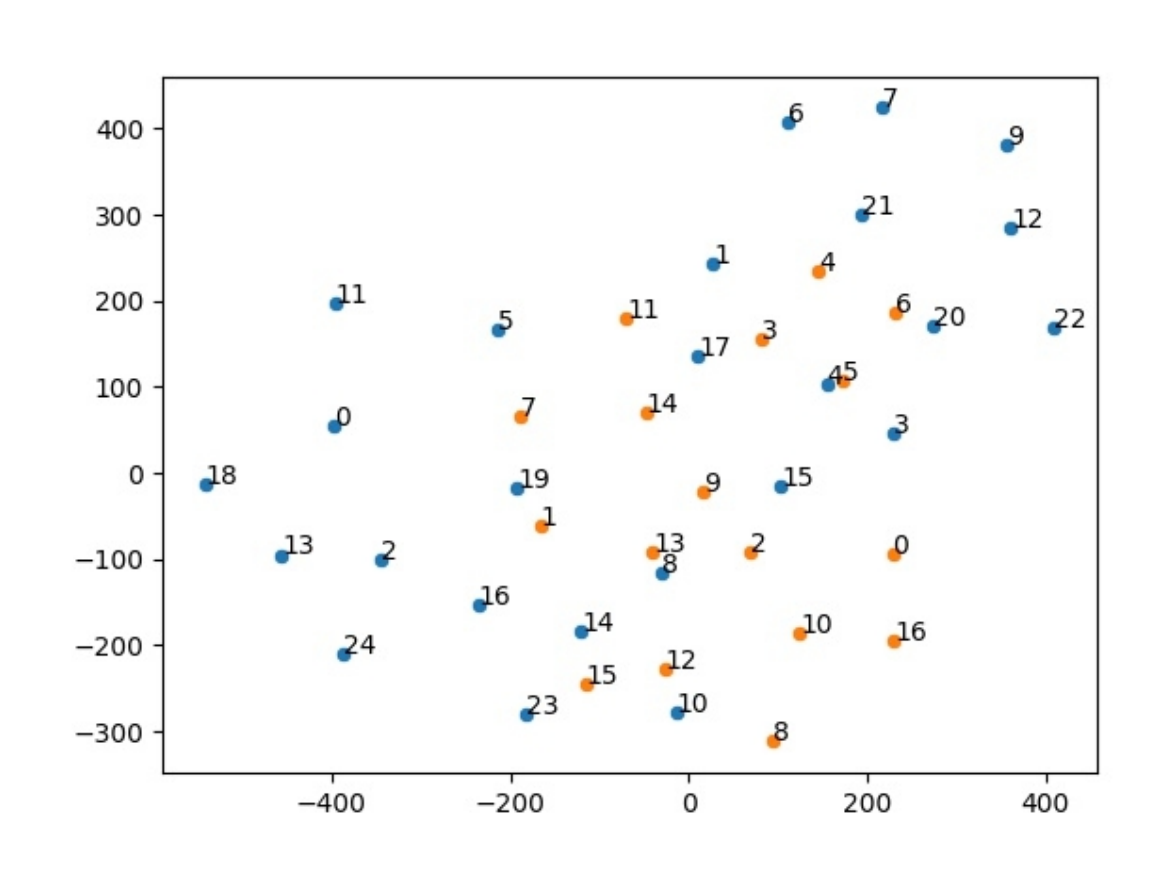}}
\caption{ The t-SNE visualisation to show embeddings of names of 17 time-series health status indicators (coloured in orange) and 25 disease risk labels (coloured in blue). Refer to Fig. \ref{fig:note}  for detailed information of these names.}
\label{fig:tsne2}
\end{figure}

Fig.~\ref{fig:tsne2} visualizes the embeddings of the names of health indicators and names of the disease risk labels in a 2D dimensional space using t-SNE, where 25 disease risk labels are coloured in blue and 17 time-series health status indicators are coloured in orange.
In Fig.~\ref{fig:tsne2}, we can see that some blue points are closely located with some orange points. For example, blue point `4' and orange `5' are close neighbours in the plot. Blue point `4' refers to `Fluid and electrolyte disorders', which is clinically related to orange point `5' (`Glasgow coma scale total') \cite{waterhouse2005glasgow}. Another example is that blue point `8' and orange point `13' are located closely. Blue point `8' is a risk label referring to 'Pleurisy; pneumothorax; pulmonary collapse', which has been found to affect orange point `13' ('Systolic blood pressure') \cite{sahn2000spontaneous}.
These observations indicate that clinically relevant medical terms tend to have similar embeddings.

\subsubsection{Interpretability of Attention}
In this work, one of the most important contributions is the high interpretability, and explainability of the results attained. Although the AI techniques, combined with extensive EHRs data have the potential to bring dramatic changes to the healthcare industry, lack of interpretability may become the bottleneck. In order for them to be readily adopted by real-world clinical practices, interpretability is crucial in medical AI without sacrificing prediction accuracy.  It will help medical professionals to decide whether they should follow a prediction or not, and meanwhile will facilitate the deployment of such systems.
In this subsection, let us focus on investigating which words/phrases in medical notes have been found to imply the risk of diseases. Results from our model are compared with those from baseline methods LEAM and $\mathcal{B}$+LEAM.

The cross-attention score is used to highlight those risk-related words. 
In this experiment, we regard the top 50\% of tokens as important ones. 

We randomly select 3 patients who have different types of disease risk. The interpretable results are shown in Table \ref{table:3Patient}, where important words from different models are bolded. 
 For patient 1, `erythema granularity friability', `stomach compatible', `gastritis erythema and congestion', and `duodenal bulb' are some explicit phrases that are related to the target disease risk labels (i.e., `Fluid and electrolyte disorders' and `Gastrointestinal hemorrhage').
 We can notice that only LDAM has highlighted  all these phrases, while LEAM and $\mathcal{B}$+LEAM have some missing. 
 Similarly, for patients 2 and 3 we can also find that 
 LDAM has generated the most interpretable results.

\subsection{Sensitivity Analysis}

\begin{figure}[htbp]
\begin{equation*}
\end{equation*}
\includegraphics[scale=0.35]{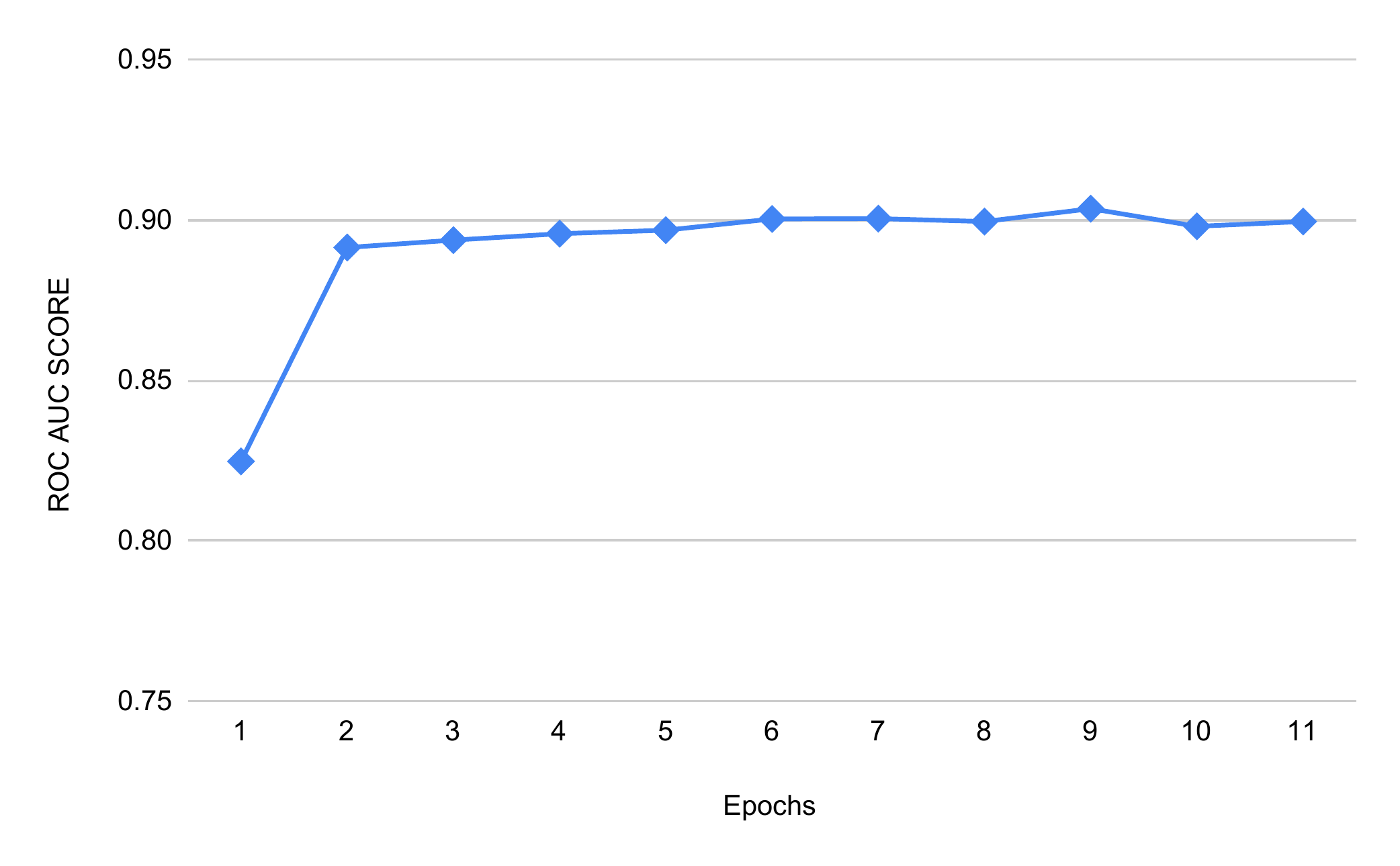}
\caption{The changes of micro ROC AUC values during training process.}
\label{fig:trainp}
\end{figure}

\begin{figure}[htbp]
\centerline{\includegraphics[scale=0.35]{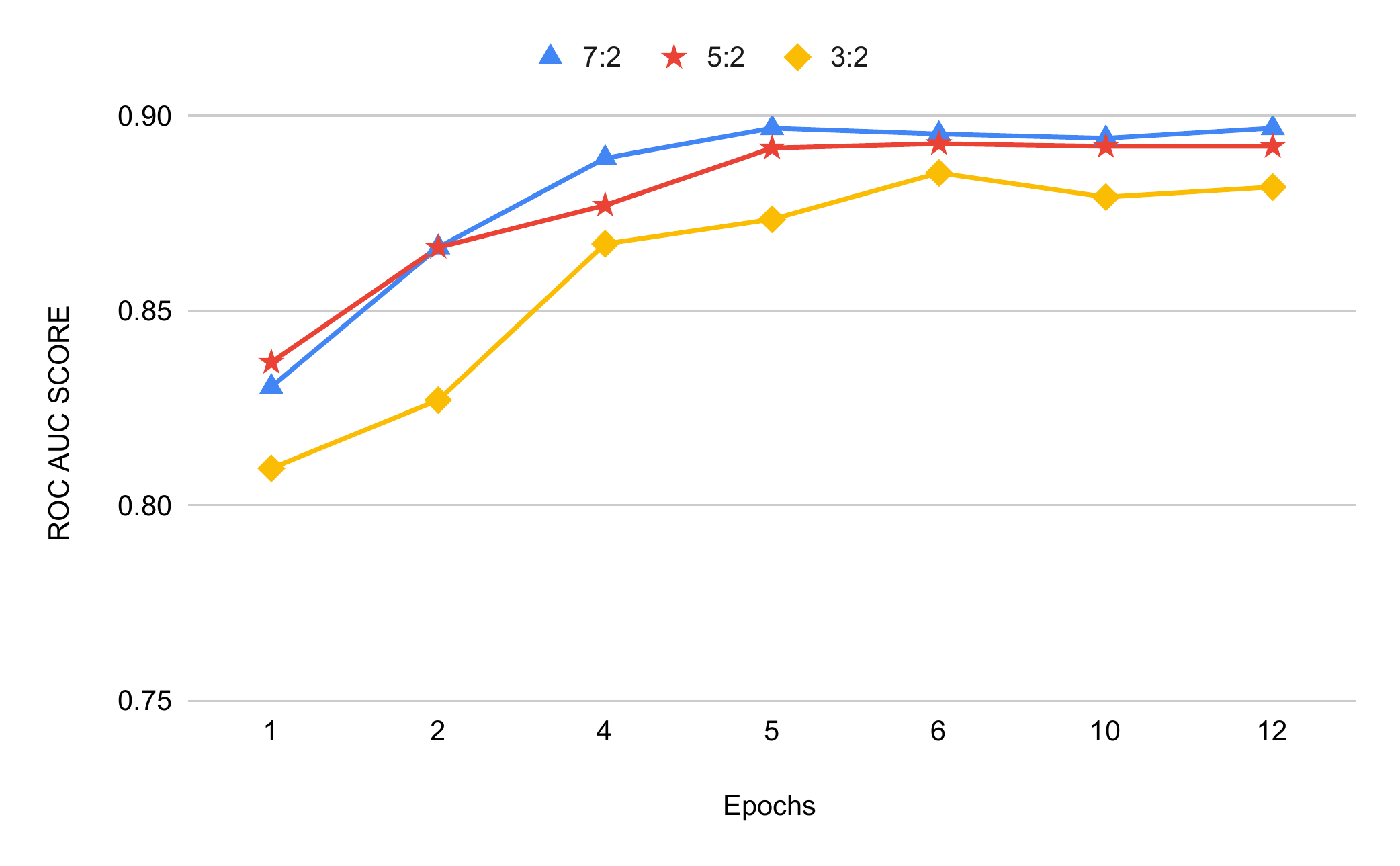}}
\caption{The Micro ROC AUC curves 
resulted from different training-to-testing ratios.}
\label{fig:ratio}
\end{figure}

\begin{figure}[htbp]
\centerline{\includegraphics[scale=0.35]{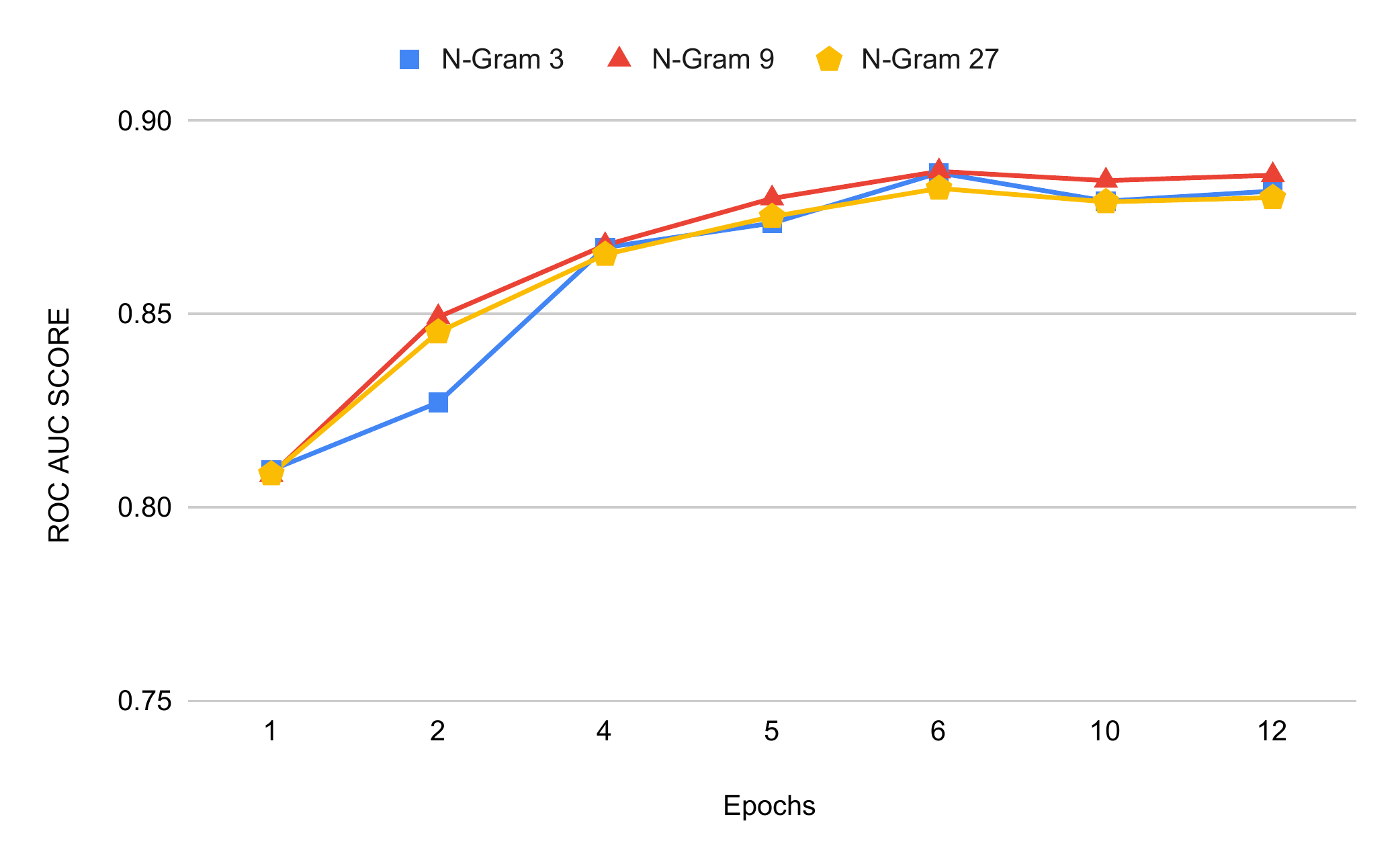}}
\caption{The Micro ROC AUC curves resulted from different values of the N-Gram size.}
\label{fig:ngram}
\end{figure}

In this section, we investigate the influence of different hyperparameters on the micro ROC AUC score.  Fig.~\ref{fig:trainp} shows the AUC scores for  different epochs during the training process. We can see that there is a sharp increase in AUC scores when the number of epochs increases from 1 to 2. Afterwards the AUC scores remain nearly stable around 0.9. This observation indicates that our model can converge quickly.
Fig.~\ref{fig:ratio} further investigates the influence of training data size. With the fixed testing dataset, we construct a training dataset based on different training-to-testing ratios, which are \{7:2,5:2,3:2\}. From Fig.~\ref{fig:ratio}, we can see that  a larger ratio would give a slightly better performance. However, the increase in the training-to-testing ratio would not greatly influence AUC, implying that our model is not sensitive to different data splitting settings. 
Fig.~\ref{fig:ngram} also shows the influence of the N-Gram size of CNN on the dataset with the training-to-testing ratios equal to 3:2. We can see that with different settings of the N-Gram size \{3,9,27\}, the AUC values show very small changes during the training process across all epochs. 


\section{Conclusions}
In this paper, we propose a label dependent attention model, namely LDAM, to predict disease risks using both time-series health status indicators, and medical notes from EHRs. To increase the interpretability of our model, we adopt a joint label and feature embedding approach and only focus on features relevant to our prediction tasks. The names of health status indicators, the names of disease risk labels, and medical notes are embedded by Clinical-BERT.  Clinical-BERT is a large-scale biomedical language model that has been pre-trained on a large clinical corpus. We have shown that using Clinical-BERT can capture the semantic meaning of features and help to generate a cross-attention vector for weighted feature integration. We have demonstrated our model on a real-world EHRs dataset. Notably, our model LDAM can generate interpretable results while achieving a state-of-the-art predictive performance. 
In the future, we can extend it to other modalities of EHRs and further apply our approach to other downstream tasks, such as drug recommendation and disease diagnosis.

{\small

\bibliographystyle{IEEEtranS}
}

\end{document}